\documentclass[a4paper,fleqn]{cas-dc}

\usepackage[numbers]{natbib}
\usepackage{tabularx}
\usepackage{bbding}
\usepackage{makecell}
\usepackage{array}
\usepackage{multirow}
\usepackage{hyperref}
\hypersetup{
    colorlinks=true,
    linkcolor=blue,
    urlcolor=blue,
    anchorcolor=blue,
    citecolor=blue
}

\usepackage{float}
\usepackage{tikz}
\usetikzlibrary{backgrounds,mindmap}
\usepackage{pgfplots}
\usepackage{listings}
\usepackage{soul}
\usepackage{color, xcolor}  
\sethlcolor{yellow}

\usepackage[switch]{lineno}   

\begin{document}

\shorttitle{Graph-Based Fraud Detection with Dual-Path Graph Filtering} 
\shortauthors{Wei He \textit{et al.}}

\title [mode = title]{Graph-Based Fraud Detection with Dual-Path Graph Filtering}              

\author[1]{Wei He}
\ead{hewei2151@gmail.com}

\author[1]{Wensheng Gan}
\cormark[1]
\cortext[cor1]{Corresponding author}
\ead{wsgan001@gmail.com}

\author[2]{Philip S. Yu}
\ead{psyu@uic.edu}

\address[1]{School of Intelligent Systems Science and Engineering, Jinan University, Zhuhai 519070, China}
\address[2]{Department of Computer Science, University of Illinois Chicago, Chicago 60607, USA}

\begin{abstract}
    Fraud detection on graph data can be viewed as a demanding task that requires distinguishing between different types of nodes. Because graph neural networks (GNNs) are naturally suited for processing information encoded in graph form through their message-passing operations, methods based on GNN models have increasingly attracted attention in the fraud detection domain. However, fraud graphs inherently exhibit relation camouflage, high heterophily, and class imbalance, causing most GNNs to underperform in fraud detection tasks. To address these challenges, this paper proposes a Graph-Based Fraud Detection Model with Dual-Path Graph Filtering (DPF-GFD). DPF-GFD first applies a beta wavelet-based operator to the original graph to capture key structural patterns. It then constructs a similarity graph from distance-based node representations and applies an improved low-pass filter. The embeddings from the original and similarity graphs are fused through supervised representation learning to obtain node features, which are finally used by an ensemble tree model to assess the fraud risk of unlabeled nodes. Unlike existing single-graph smoothing approaches, DPF-GFD introduces a frequency-complementary dual-path filtering paradigm tailored for fraud detection, explicitly decoupling structural anomaly modeling and feature similarity modeling. This design enables more discriminative and stable node representations in highly heterophilous and imbalanced fraud graphs. Comprehensive experiments on four real-world financial fraud detection datasets demonstrate the effectiveness of our proposed method. The source code is publicly available at \url{https://github.com/vidahee/DPF-GFD}.
\end{abstract}


\maketitle

\section{Introduction}  \label{sec: introduction}

In recent years, the global digital economy has demonstrated robust growth, with the integration of digital technology and the industrial economy continuously deepening. Concurrently, novel financial crime tactics and types have emerged, among which financial fraud \cite{reurink2016financial} has become a pervasive global threat. Financial fraud not only inflicts substantial economic losses on individuals and enterprises but also undermines the stability of domestic and global economies by eroding public trust in financial systems and compromising their integrity \cite{cheng2025graph}. Generally, fraud encompasses both individual fraud and organized fraud. With the advancement of anti-fraud techniques and technologies, fraudulent activities have become increasingly difficult for individuals to execute independently; instead, they rely on organized fraud rings. Currently, the primary challenges in financial anti-fraud are twofold: First, the accounts, devices, and environmental factors involved in financial operations exhibit extensive interconnections with complex business logic. The factors associated with certain fraudsters are relatively concealed, manifesting signals that are often not directly observable and require indirect methods for perception and understanding. Second, from traditional hacking attacks, phishing websites, and Trojan viruses to emerging deepfakes and biometric information theft, fraudsters continuously evolve their tactics to evade regulation and circumvent systems. This leads to constantly changing risk patterns and emerging risk types, making it difficult for financial institutions to identify and prevent fraud.

Financial fraud detection systems aim to identify fraudulent activities within systems and report them to system administrators through various techniques \cite{motie2024financial}. Early fraud detection solutions largely relied on manually crafted rules \cite{bolton2002statistical} and conventional machine-learning models \cite{hernandez2024financial}. However, they exhibit limitations in handling increasingly complex fraud patterns and large-scale data. Deep learning has therefore gained prominence in the financial-fraud domain, as its representation-learning capabilities enable the extraction of intricate patterns directly from raw inputs \cite{li2022internet}. In recent years, graph-based deep learning has advanced rapidly, with numerous innovative graph neural networks (GNNs) \cite{wu2020comprehensive} being proposed and applied across various domains. Given their capacity to model complex relational dependencies, GNNs have found broad utility in financial applications, such as credit card fraud analytics, taxation-related anomaly detection, cryptocurrency fraud surveillance, insurance-claim verification, online-payment integrity checks, and other high-risk transactional environments. For instance, the CARE-GNN model \cite{dou2020enhancing} effectively enhances GNN robustness against camouflaged fraudsters through a multi-channel aggregation mechanism based on behavioral similarity and neighbor credibility. ASA-GNN \cite{tian2023asa} balances model efficiency and accuracy by leveraging adaptive sampling and aggregation strategies, thereby better capturing abnormal transaction patterns. GCD-GNN \cite{liu2025global} constructs a global confidence-based graph neural network that improves fraud identification stability and generalization by incorporating nodes' global confidence information. Most recently, CAFD \cite{lou2025graph} achieves superior performance in transaction fraud detection tasks by effectively integrating nodes' local structural and global semantic information within a context encoding and adaptive aggregation framework. Overall, these studies have continuously advanced GNN development in financial fraud detection, demonstrating significant advantages in identifying complex relationship networks and concealed anomalous behaviors.

The emergence of GNNs effectively addresses the limitation of ignoring user interactions. However, current GNN-based financial fraud detection methods face challenges and have not produced satisfactory results. First, financial fraud scenarios often encounter the relation camouflage problem \cite{dou2020enhancing}. Real-world fraudsters frequently establish false relationships with numerous benign entities while hiding direct relationships with other fraudsters. Benign neighborhood information with relation camouflage reduces the suspiciousness of fraud nodes, thereby undermining discriminative anomaly information. Second, many GNN architectures are grounded in the homophily principle, in which information from nearby nodes is combined under the expectation of label consistency, thereby making the resulting propagation behavior similar to that of a low-pass filter \cite{wang2022powerful}. When a graph instead exhibits heterophily and links frequently connect nodes from different categories, this smoothing effect can suppress subtle but informative irregularities. Finally, because fraudulent instances usually represent only a very small portion of the entire dataset, a pronounced data imbalance arises, and this severely impacts the stability and performance of existing methods.

To address these challenges, we propose a novel financial fraud detection method called the Graph-based Fraud Detection model with Dual-Path graph Filtering (DPF-GFD). DPF-GFD comprises several key modules designed to tackle relation camouflage, high heterophily, and class imbalance challenges, thereby improving financial fraud detection accuracy. Specifically, the proposed framework begins by employing a Beta wavelet-based operator on the original graph to extract structural information, as this filter flexibly responds to different signal-frequency components. Next, a similarity graph is constructed by computing Euclidean-distance relationships among the node feature vectors, and an enhanced all-pass filtering module is applied to this graph, enabling behavior comparable to low-pass smoothing when similar feature patterns need to be emphasized. The representations derived from both the original graph and the similarity graph are then integrated through a supervised representation-learning module to generate refined node embeddings. Finally, these embeddings are passed to an ensemble tree model to estimate the fraud likelihood of unlabeled nodes. Our main contributions can be summarized as follows:

\begin{itemize}
    \item We introduce a fraud-oriented dual-path graph filtering paradigm that explicitly disentangles structural anomaly modeling from feature-consistency reconstruction. Unlike conventional graph-based fraud detection methods that rely on a single low-pass propagation mechanism, the proposed framework coordinates two complementary graph views to better handle relation camouflage and heterophily in fraud graphs.
    \item We propose a frequency-complementary filtering strategy tailored to fraud detection. A constrained Beta wavelet operator selectively enhances mid- to high-frequency structural irregularities on the original transaction graph, while a similarity-graph low-pass filter stabilizes feature propagation under heterophily. This coordinated design enables controlled anomaly amplification without over-smoothing.
    \item We conduct extensive experiments and controlled studies demonstrating that explicitly modeling multi-frequency graph signals leads to more discriminative and stable fraud representations across datasets with diverse graph structures and imbalance levels.
\end{itemize}

The remainder of this paper is organized as follows: Section \ref{sec: introduction} provides an overall introduction. Section \ref{sec: relatedwork} reviews related work on financial fraud detection. Section \ref{sec: Problem Definition} introduces the notation used in this paper and formally defines the research problem. Section \ref{sec: DPF-GFD Method} provides an in-depth description of the DPF-GFD method. Section \ref{sec: Experiments} presents the datasets and comparison methods, along with experimental results. Section \ref{sec: conclusion} concludes this work and outlines future research directions.

\section{Related Work} \label{sec: relatedwork}

Over the past several decades, extensive studies have been conducted to explore fraud detection and prevention in finance. Financial fraud detection primarily comprises expert system-based methods, traditional machine learning-based methods, and deep learning-based methods.

\subsection{Expert System-Based Methods}

These methods primarily leverage domain experts' experience and knowledge, employing rule-based and expert-strategy approaches to risk control. In early stages, data used for fraud detection was typically highly structured, and expert systems commonly employed symbolic rules to encode knowledge created by human experts. These rules and static thresholds could be used to filter inappropriate behaviors \cite{zhu2021intelligent}. Vorobyev \textit{et al.} \cite{vorobyev2022reducing} proposed an automatic rule generation framework for bank anti-fraud that implements tree-based machine learning algorithms to generate new rules with features composed of existing expert rule components, thereby identifying misclassified fraud cases. This method improved rule-generation automation to some extent but still relied on manually designed features and thresholds, making it difficult to address rapidly evolving fraud patterns. Martins \textit{et al.} \cite{martins2024riff} proposed the RIFF method, which emphasizes deriving concise rule sets from single or multiple decision trees that remain robust under low false positive rates, aiming to provide auditors with few, precise, and highly interpretable rules. However, this method has limited applicability to high-dimensional data or nonlinear fraud patterns and lacks adaptability when facing diverse transaction structures. Overall, rule-based systems operate on fixed rules and thresholds, such as flagging transactions exceeding specific amounts or originating from high-risk regions as fraudulent. Although these methods are easy to deploy and understand, they are inherently passive defenses unable to adapt to emerging fraud strategies \cite{manzoor2025enhancing}. Fok \textit{et al.} \cite{fok2025foe} demonstrated through transferable adversarial attack experiments that merely applying slight perturbations to input features enables fraud samples to cross detection boundaries, thereby bypassing models and rule systems. This result reveals fundamental limitations of traditional rule sets when facing adaptive fraud strategies: lagging rule updates, poor generalization, and easily exploitable boundary definitions. As fraud tactics and model attack techniques continue evolving, rule-based expert systems face enormous challenges in accuracy, scalability, and maintenance costs, urgently requiring detection mechanisms with stronger self-learning and generalization capabilities.

\subsection{Traditional Machine Learning-Based Methods}

These methods often rely on manual feature engineering, including feature extraction, selection, and construction. Feature engineering quality directly impacts machine learning model performance and prediction results. As the limitations of expert systems in complex fraud scenarios became apparent, researchers began employing traditional machine learning algorithms to model financial data for more efficient fraud detection. Deng \textit{et al.} \cite{deng2010detection} adopted Naïve Bayes classifiers to establish conditional probability relationships between financial indicators and fraud labels, achieving automatic identification of financial statement anomalies and validating the feasibility of statistically-based models in fraud detection. However, this method relies on conditional independence assumptions among features, which can lead to decreased accuracy in high-dimensional, strongly correlated real-world data. Sahin \textit{et al.} \cite{csahin2011detecting} investigated performance differences between decision trees and support vector machines (SVM) in credit card fraud identification, revealing the strengths and weaknesses of different models in feature selection, nonlinear relationship modeling, and class imbalance handling. However, both methods are sensitive to sample distributions and prone to overfitting or high false positive rates on extremely imbalanced datasets. Liu \textit{et al.} \cite{liu2015financial} utilized Random Forest algorithms, effectively improving detection robustness and accuracy through ensemble learning with random feature sampling and voting mechanisms across multiple decision trees. Nevertheless, Random Forest models have complex parameters, poor interpretability, and may experience performance fluctuations with high feature noise. Gyamfi \textit{et al.} \cite{gyamfi2018bank} applied SVM to bank transaction data, enhancing model discrimination of complex transaction patterns through kernel function mapping while emphasizing feature engineering's critical role in fraud identification. Although SVM demonstrates good performance on small- to medium-sized datasets, it incurs high computational costs when processing large-scale, dynamically updated transaction streams and struggles with online model updates as data changes. Mishra \textit{et al.} \cite{mishra2021fraud} introduced logistic regression with k-fold cross-validation to improve model generalization and stability, further validating traditional supervised learning algorithms' applicability across diverse fraud scenarios. However, logistic regression strongly assumes linear separability of features, making it difficult to capture complex nonlinear fraud relationships.
In summary, these studies demonstrate extensive exploration of traditional machine learning models in financial fraud detection. Compared to static rule systems, they can automatically extract latent features from data and adaptively update detection patterns. However, these models generally depend on high-quality feature engineering and have limited capability in modeling nonlinear patterns and temporal features, laying the foundation for introducing deep learning methods.

\subsection{Deep Learning-Based Methods}

These methods can extract valuable information from massive data, thereby achieving highly automated and intelligent fraud detection. As traditional machine learning models' limitations in high-dimensional nonlinear data modeling gradually became apparent, researchers began introducing deep learning techniques into financial fraud detection to automatically extract complex features and improve model adaptability. Gómez \textit{et al.} \cite{gomez2018end} proposed an end-to-end Artificial Neural Network (ANN) architecture that models credit card transaction data through multilayer perceptron structures, achieving end-to-end automation from feature learning to fraud scoring while significantly reducing dependence on manual feature engineering. However, due to ANN's relatively simple structure, it has limited capability in capturing temporal dependencies and local feature correlations and is prone to overfitting. Fu \textit{et al.} \cite{fu2016credit} applied Convolutional Neural Networks (CNN) to transaction data, utilizing local receptive field mechanisms to capture latent correlations in feature space, thereby improving identification of abnormal transaction patterns. However, CNN models are better suited for features with spatial correlations and have relatively limited capability in modeling high-dimensional interactive features and cross-entity associations in financial transaction data. Zhang \textit{et al.} \cite{zhang2018model} proposed an improved CNN model with multi-scale convolutional layers to enhance complex transaction feature representation, further improving model accuracy and generalization in anomaly detection tasks. However, this method is sensitive to hyperparameter settings, and as network depth increases, model complexity and computational costs rise significantly, making it difficult to meet real-time detection requirements. Branco \textit{et al.} \cite{branco2020interleaved} introduced Gated Recurrent Unit (GRU) structures to dynamically model transaction sequences, effectively capturing temporal dependencies in user behavior and long-term feature evolution. Although RNN-based models excel in temporal modeling, their training process is prone to vanishing or exploding gradients and has low efficiency in modeling long sequences. Overall, deep learning methods have achieved substantial breakthroughs in feature representation and pattern recognition compared to traditional machine learning, enabling fraud detection systems to automatically learn latent feature relationships from raw data with stronger nonlinear modeling capabilities. However, these methods generally suffer from high model complexity, insufficient interpretability, and strong dependence on large-scale labeled data, prompting researchers to further explore graph neural network methods that can combine structured relational information with contextual semantics.

Among deep learning-based fraud detection research, due to GNNs' inherent advantages in processing graph-structured data through message-passing mechanisms, GNN-based methods have gradually become a prominent research direction in fraud detection \cite{li2024research}. In the CARE-GNN model \cite{dou2020enhancing}, label-aware similarity and reinforcement-learning-based neighbor-selection mechanisms are introduced to address common feature-camouflage and relation-camouflage strategies used by fraudsters, thereby enhancing the model’s robustness against concealed fraud. However, this model incurs high computational costs when processing large-scale graph data, and the reinforcement learning module's training process is complex, limiting its real-time detection capability. Subsequently, Li \textit{et al.} \cite{li2021fraud} introduced self-attention mechanisms in their research, enabling models to dynamically allocate neighbor weights and thereby improve feature aggregation precision and representation capability. However, introducing self-attention structures significantly increases model parameters, imposing higher requirements on training data scale and hardware resources and potentially causing overfitting. In the ASA-GNN model \cite{tian2023asa}, over-smoothing and neighbor redundancy are mitigated through adaptive sampling and multi-hop neighbor aggregation, resulting in improved stability and generalization performance in fraud detection. However, this method still relies on hyperparameter tuning for neighbor sampling strategies and has limited performance on extremely sparse transaction networks. Liu \textit{et al.} \cite{liu2025global} further proposed a global confidence-based graph neural network that evaluates node typicality through prototype learning and distinguishes typical from atypical neighbors during aggregation, thereby strengthening model discrimination. However, this model still faces challenges in cross-scenario transferability across different financial contexts, and the prototype learning process is sensitive to anomalous node distributions. Recently, Lou \textit{et al.} \cite{lou2025graph} more effectively identified camouflaged and group fraud through context encoding and adaptive aggregation strategies combined with temporal interval and behavioral pattern information. Overall, these studies have advanced fraud detection's evolution from feature-level analysis to relational structure modeling, enabling models to demonstrate stronger representation and generalization capabilities in high-dimensional, dynamic, adversarial financial scenarios.

From an overall perspective, financial fraud detection technology development has progressed from expert knowledge and rule-reasoning-based expert systems, to statistical feature modeling-based traditional machine learning methods, to deep learning models capable of automatically extracting deep features, and finally to graph neural network frameworks integrating structural relationships and contextual information. This technological trajectory reflects the evolution of fraud detection research from experience-driven to data-driven and ultimately to relation-driven approaches, establishing a solid foundation for constructing more intelligent and interpretable financial risk control systems.

\section{Problem Definition} \label{sec: Problem Definition}

We first introduce the notation used in this paper, followed by a formal problem definition.

A graph is a collection of nodes and edges, denoted as $G$ = ($V$, $E$, $X$), where $V$ = \{$v_1$, $v_2$, $\cdots$, $v_N$\} is the set of nodes with $N$ denoting the number of nodes. $E$ = \{$e_1$, $e_2$, $\cdots$, $e_M$\} is the set of edges with $M$ denoting the number of edges, where edge $e$ = $(v_i,v_j)$ $\in$ $E$ represents the connection relationship between nodes $v_i$ and $v_j$ in the graph. $X \in \mathbb{R}^{N \times F}$ represents the node feature matrix, where $F$ denotes the dimensionality of node feature vectors. Each node $v_i$ is associated with an F-dimensional feature vector $x_i$ $\in \mathbb{R}^F $, and the collection of all node features is $X$ = $\{x_1,x_2,\cdots,x_N\}$. The adjacency matrix of the graph can be defined as $A \in \mathbb{R}^{N \times N}$, where $A_{ij}$ = $1$ if $(v_i,v_j) \in E$ and $A_{ij}$ = $0$ otherwise. 

Let $V_f$ and $V_b$ be two disjoint subsets of $V$, where $V_f$ denotes all nodes labeled as fraudulent, and $V_b$ denotes all nodes labeled as benign. Graph-based fraud detection involves classifying unlabeled nodes in $G$ as either fraudulent or benign, given the graph structure $E$, node features $X$, and partially labeled nodes $V_f$ and $V_b$. In most real-world settings, benign nodes constitute the vast majority of the graph, while fraudulent nodes appear only sparsely. As a result, financial fraud detection on graphs naturally corresponds to a supervised node-level task characterized by severe class imbalance. Each node $v$ is embedded into a low-dimensional representation $h_v \in \mathbb{R}^D,$ and a predictor $P(\cdot;{\theta}_p)$ is employed for prediction: 
\begin{equation}
    p_v = P(h_v;{\theta}_p)= \text{Softmax}(h_v+b_p),
\end{equation}
where $\theta_p$ = $\{W_p \in {\mathbb{R}^{K \times D}},b_p \in \mathbb{R}^K \} $ are learnable parameters with $K$ = 2 in fraud detection scenarios. For a labeled node $v$, let $y_v \in \mathbb{R}^K$ represent a one-hot label vector, where $y_v[k]$ = $1$ if and only if the node $v$ belongs to class $k$. Given a training set $V_{tr}$, the loss function is formulated by applying cross-entropy loss to predictions:
\begin{equation}
L=-\sum_{v \in {V_{tr}}}{\sum_{k=0}^{K}{y_v}}[k] \text{ln} {p_v}[k].
\end{equation}

\section{DPF-GFD Method}  \label{sec: DPF-GFD Method}

In this section, we introduce DPF-GFD, a frequency-complementary dual-path graph filtering framework tailored for fraud detection on heterophilous and relation-camouflaged graphs. Instead of relying on single-graph smoothing, DPF-GFD explicitly decouples structural anomaly modeling and feature similarity modeling through two complementary graph filtering paths. The overall architecture of DPF-GFD is illustrated in Figure \ref{fig:DPF-GFD}. The model comprises four main components: an adaptive filtering module for the original graph, a low-pass filtering module for the neighboring graph, a feature fusion module, and a final ensemble tree classification module. The adaptive filtering module captures multi-frequency information through adjustable filters while enhancing responses to high-frequency anomaly signals. The low-pass filtering module reconstructs inter-node relationships at the structural level and smooths features, effectively mitigating relation camouflage and heterophily issues. The feature fusion module concatenates outputs from both paths to form a more discriminative joint representation. The ensemble tree classification module leverages ensemble learning's discriminative advantages to achieve accurate identification and robust detection of fraud nodes. In the following subsections, we detail the main components and architectural specifics of DPF-GFD.

\begin{figure*}[htbp]
    \centering
    \includegraphics[width=1.0 \textwidth]{figures/model.pdf}
    \caption{Overview of the proposed DPF-GFD framework. The model adopts a dual-path graph filtering design to address relation camouflage, heterophily, and class imbalance in fraud graphs. The structural path enhances multi-frequency structural anomalies on the original graph, while the similarity-based path applies low-pass smoothing on a kNN graph to reconstruct feature consistency. The fused representations are then fed into an imbalance-robust ensemble classifier for fraud risk prediction.}
    \label{fig:DPF-GFD}
\end{figure*}

\subsection{Adaptive Filtering Module}

In spectral graph neural network research, BernNet \cite{he2021bernnet} achieves arbitrary spectral filter learning through Bernstein polynomial approximation, flexibly allocating weights across different frequency components. Specformer \cite{bo2023specformer} further introduces Transformer structures to model dependencies between frequency components in the spectral domain, thereby exhibiting stronger filtering representation capabilities. Although these methods perform well in general node classification tasks, their performance is often limited in scenarios with severe class imbalance, such as financial fraud detection. Since fraud nodes constitute only a small proportion of the graph, overall graph signals exhibit pronounced low-frequency dominance. During model training, filters more easily develop preferences for low-frequency components, leading to over-smoothing or neglect of high-frequency information corresponding to anomalous nodes. The Beta Wavelet-based operator \cite{tang2022rethinking} is employed as a band-pass filter to adaptively capture specific and appropriate signal frequencies in the spectral domain, enabling the enhancement of high-frequency anomalous signals while preserving low-frequency smooth features.

This subsection first defines the relevant mathematical foundations. Specifically, spectral graph filtering relies on the spectral decomposition of the graph Laplacian matrix, necessitating a clear definition of the normalized graph Laplacian matrix. Subsequently, to construct band-pass filters with adjustable frequency responses, we introduce the Beta distribution with flexible shape parameters as the mathematical foundation for the filter kernels. Finally, based on this distribution function, we adopt the Beta Wavelet-based operator \cite{tang2022rethinking} to adaptively extract structural information from the original graph.

First, given an undirected graph $G$ = $(V,E,X)$ with an adjacency matrix $A$ and a degree matrix $D$, according to spectral graph theory, the normalized graph Laplacian matrix \cite{chung1997spectral} is defined as:
\begin{equation}
L = I - D^{-\frac{1}{2}}AD^{-\frac{1}{2}},
\end{equation}
where $I$ is the identity matrix. Since the graph $G$ is undirected, $L$ is a real symmetric positive semi-definite matrix, guaranteeing the eigen decomposition of the graph Laplacian: $L$ = $U \lambda U^T$, where $ U=[u_0,u_1,\cdots,u_{n-1}] $ represents the eigenvectors of the Laplacian matrix, and $\lambda$ = $\text{diag}( {\lambda}_0, {\lambda}_1,\cdots, {\lambda}_{n-1})$ is the diagonal matrix of eigenvalues. Through eigen decomposition of $L$, the spectral representation of graph signals can be obtained to describe signal variations across different frequency components.

Second, to design filter functions that can flexibly adjust their shapes across different frequency ranges, we introduce the Beta distribution as the mathematical basis for the filter kernels. The Beta distribution is a continuous probability distribution defined in $[0,1]$, with its corresponding probability density function (PDF) \cite{johnson1994beta} given by:
\begin{equation}
\beta_{p,q}(w) =
\begin{cases}
\dfrac{1}{B(p+1, q+1)} w^p (1 - w)^q, & \text{if } w \in [0, 1] \\[4pt]
0, & \text{otherwise},
\end{cases}
\end{equation}
where $p,q \in \mathbb{R}^+$, and $B(p+1,q+1)$ = $p!q!/(p+q+1)!$ is a constant. $p$ and $q$ are shape parameters controlling energy distribution in low-frequency and high-frequency regions, respectively. By adjusting $p$ and $q$, different filter shapes and bandwidth control can be achieved.

\begin{figure}
    \centering
    \includegraphics[width=0.45\textwidth]{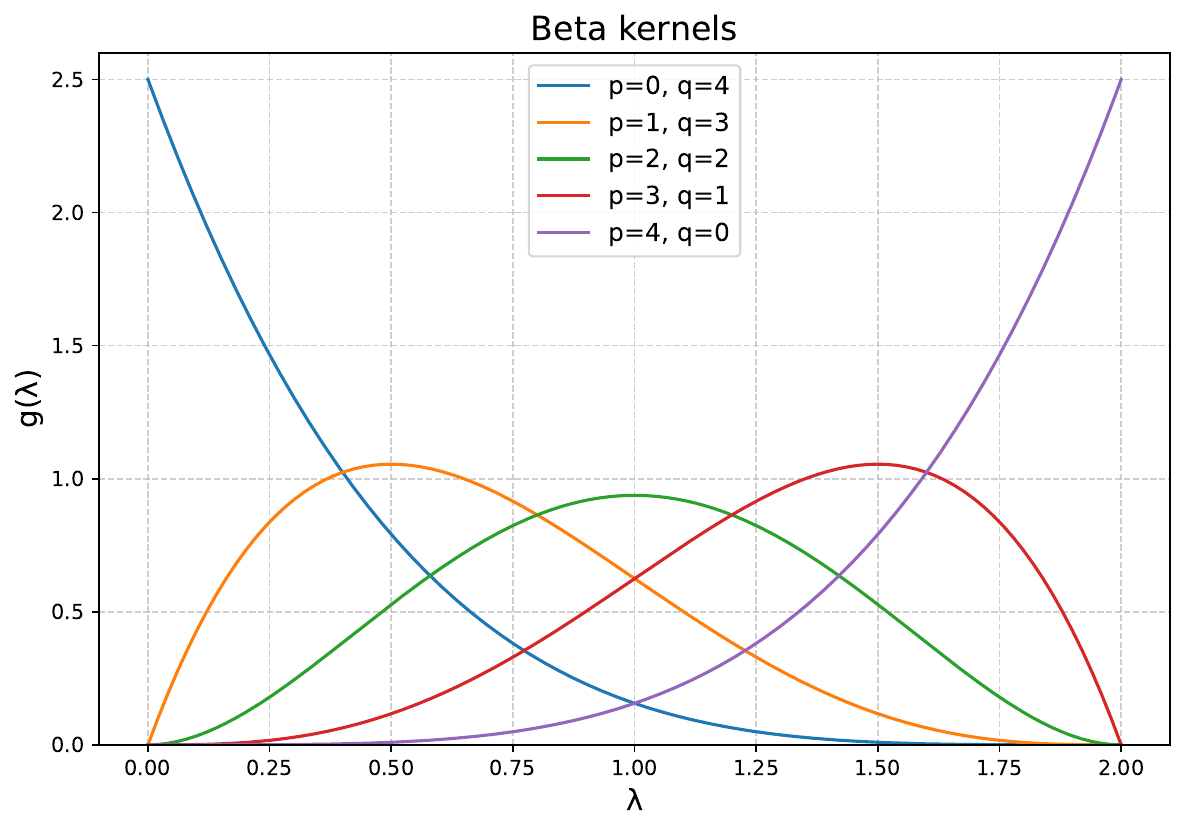}
    \caption{Beta wavelet-based kernels in the spectral domain when $C = 4$.}
    \label{fig:beta_kernels}
\end{figure}

Since the eigenvalues of the normalized graph Laplacian $L$ satisfy $\lambda \in [0,2]$, we use ${\beta}^{*}_ {p,q}(w)=\frac{1}{2}{
\beta_{p,q}}(\frac{w}{2})$ to cover $L$'s entire spectral range and further add the constraint $p,q \in \mathbb{N}^+$ to ensure $\beta^{*}(p,q)$ is a polynomial. Therefore, the Beta Wavelet-based operator $W_{p,q}$ \cite{tang2022rethinking} can be written as:
\begin{equation}
    	{W_{p,q}}=U{\beta ^{*}_{p,q}}(\lambda)U^T={\beta ^{*}_{p,q}}(L)=\frac{(L/2)^p(I-L/2)^q}{2B(p+1,q+1)}.
\end{equation}
Let $p+q$ = $C$ be constant; the Beta Wavelet-based operator $W$ comprises a set of $C+1$ Beta kernels with the same order:
\begin{equation}
W = (W_{0,C}, W_{1,C-1}, \cdots, W_{C,0}),
\label{eq:beta wavelet}
\end{equation}
In this equation, $W_{0,C}$ functions as a kernel emphasizing low-frequency components, while $W_{C,0}$ emphasizes high-frequency ones; the remaining configurations behave as band-pass operators with varying spectral widths. Assuming $p+q$ = 4, corresponding to $C$ = 4, there are five cases: $(p,q)$ = (0,4), (1,3), (2,2), (3,1), (4,0). Visualizing the range of $\beta ^*_{p,q}(w)$ as shown in Figure \ref{fig:beta_kernels}, we observe that when $p$ = 0, $\beta^*_{p,q}(w)$ shows its strongest response in the low-frequency region, whereas when $q$ = 0, its dominant response shifts toward the high-frequency region, showing responses comparable to operators focusing on either low-frequency or high-frequency bands. Examining the beta kernels further reveals that varying the pair $(p,q)$ modulates the spectral focus of the operator, enabling sensitivity to different portions of the frequency spectrum.

Based on this principle, for the original graph, we employ the Beta Wavelet-based operator as the core spectral filtering operator to adaptively extract structural information, with the mathematical form:
\begin{equation}
Z_{1i} = W_{i,C-i}(X) .
\label{eq:adaptive representation}
\end{equation}

Through this adjustable band-pass filtering mechanism, the model can more effectively capture multi-scale associations and latent anomaly patterns among nodes in the graph, thereby improving the identification of fraudulent behaviors in complex financial transaction networks.

\subsection{Low-Pass Filtering Module}

In real financial transaction networks, fraud nodes often establish false or anomalous relationships with numerous benign nodes to conceal their behavioral characteristics, creating the relation camouflage phenomenon. Additionally, some transaction graphs exhibit significant heterophily characteristics, where connected nodes have substantial differences in semantic attributes or labels, weakening traditional GNNs' feature propagation effectiveness based on homophily assumptions. To address these issues, this subsection introduces a low-pass filtering module based on a k-Nearest Neighbor graph (kNN-Graph) following the spectral filtering module to reconstruct the graph structure and perform smoothing optimization in the feature space.

k-Nearest Neighbor graphs have been frequently studied for capturing nodes' underlying structure in feature space, effectively overcoming erroneous connections of anomalous nodes \cite{li2019spam}. The construction of the kNN-based similarity graph is driven solely by feature information rather than the original topology. Consequently, attempts to disguise fraudulent vertices through structural manipulation become largely ineffective \cite{li2025universal}. Specifically, constructing a kNN-Graph first requires calculating the distance matrix $D \in \mathbb{R}^{N \times N}$ among $N$ points. There are many methods to obtain $D$, and the DPF-GFD model employs two popular approaches: the Euclidean distance and the cosine distance. The Euclidean distance calculation formula is:
\begin{equation}
{ed}_{i,j} =||x_i-x_j|| ,
\label{eq:euclidean dist}
\end{equation}
where $x_i$ is the feature vector of node $i$, $||x_i||$ is the Euclidean norm of vector $x_i$, and ${ed}_{i,j}$ is the distance between nodes $i$ and $j$. For cosine distance, cosine similarity is first defined as follows:
\begin{equation}
{s}_{i,j}=\frac{x_i \cdot x_j}{||x_i|| \cdot ||x_j||}.
\label{eq:cosine sim}
\end{equation}
The cosine distance is then defined as:
\begin{equation}
{cd}_{i,j}=1-{s}_{i,j}.
\label{eq:cosine dist}
\end{equation}
For each node, the model identifies the k nearest nodes from the distance matrix to obtain the adjacency matrix $A_\text{kNN}$ of the kNN-Graph:
\begin{equation}
A_{\text{kNN}}(i,j) =
\begin{cases}
d_{ij}, & \text{if } j \in \text{Top-}k(i) \\[4pt]
0, & \text{otherwise},
\end{cases}
\label{eq:knn_adj}
\end{equation}
where $d_{i,j}$represents the distance between nodes $i$ and $j$, namely ${ed}_{i,j}$ in Eq.~\eqref{eq:euclidean dist} or ${cd}_{i,j}$ in Eq.~\eqref{eq:cosine dist}. The model constructs a graph based on an adjacency matrix $A_\text{kNN}$. This structural reconstruction approach does not depend on the original transaction graph's edge relationships but generates connections based on node similarity in feature space, thereby reducing the possibility of fraud nodes achieving camouflage through fabricated edges while mitigating feature mixing issues caused by heterophily.

After obtaining the kNN-Graph, we further apply a low-pass filter to smooth node features and suppress noisy edge effects. For the constructed neighbor graph, we use the low-pass filtering component from Eq. \eqref{eq:beta wavelet} to adaptively capture representations of the similarity graph.

\subsection{Feature Fusion Module}

The aforementioned two modules model node features from different perspectives: the adaptive filtering module captures multi-frequency information through adjustable filters in the spectral domain, enhancing the response to high-frequency anomalous signals; the low-pass filtering module, based on a k-nearest neighbor graph, reconstructs inter-node relationships and smooths features at the structural level, effectively alleviating the problems of relation camouflage and heterophily. To fully leverage the complementarity of these two representations, we concatenate their output features to form a more discriminative joint representation. $Z_{1i} \in \mathbb{R}^{N \times F_1}$ in Eq.~\eqref{eq:adaptive  representation} represents the node representation after adaptive filtering, and $Z_{2i} \in  \mathbb{R}^{N \times F_2}$ in Eq.~\eqref{eq:node  representation} represents the node representation after k-Nearest Neighbor graph low-pass filtering. The fused features can be expressed as:
\begin{equation}
    Z=[Z_{1i} || Z_{2i}],
\label{eq:node representation}    
\end{equation}
where $[\cdot||\cdot]$ denotes the feature concatenation operation. The fusion operation is implemented via feature concatenation to preserve the complementary information captured by the two graph paths. The original graph path focuses on structural anomaly modeling through frequency-selective filtering, while the similarity graph path emphasizes feature-level consistency via low-pass smoothing. Concatenation retains the full representational capacity of both embeddings without imposing prior assumptions on their relative importance. Moreover, since the downstream classifier is XGBoost, which inherently performs nonlinear feature selection and importance weighting, the model can automatically learn how to balance the contributions of the two paths.

Subsequently, the fused features are fed into a Multi-Layer Perceptron (MLP) \cite{goodfellow2016deep} for nonlinear mapping to obtain the final node embedding representation. Through its stacked structure of multiple linear transformations and nonlinear activations, the MLP can achieve more complex nonlinear mappings in the feature space, effectively fusing multi-scale information from both spectral and structural domains. On the one hand, it enhances the model's expressive power, enabling the model to capture potential high-order interaction relationships among node features; on the other hand, the nonlinear activation functions introduce feature sparsity and nonlinear decision boundaries, thereby improving the separability and robustness of downstream classifiers. Let the fused feature be $ Z \in \mathbb{R}^{N \times (F_1 + F_2)}$, then the computation process of the MLP can be expressed as:
\begin{equation}
    H = \text{MLP}(Z) = \sigma (W_2 \cdot \sigma(W_1 Z + b_1) + b_2),
\end{equation}
where $W_1$, $W_2$, $b_1$, and $b_2$ are learnable parameters of the MLP, and $\sigma(\cdot) $ represents the nonlinear activation function.

The generated node embeddings not only preserve multi-scale feature responses in the frequency spectrum but also possess smoother and more discriminative feature representations at the structural level, providing high-quality input features for subsequent ensemble tree classifiers.

\subsection{Ensemble Tree Classification Module}

The aforementioned modules obtain node embedding representations $H$ through joint modeling of spectral and structural domain features, which retain multi-frequency information while integrating local structural relationships. To further accomplish fraud node identification, we feed the embeddings into an ensemble tree algorithm for classification. Tree-based ensembles have proven highly competitive for detecting irregular patterns in structured datasets. Given the potential of ensemble trees, some studies have explored their application in graphs to simultaneously utilize structural and feature information, demonstrating that tree ensemble models can potentially outperform GNNs when combined with simple structural neighborhood aggregation \cite{tang2023gadbench}. XGBoost \cite{chen2016xgboost} is an ensemble learning algorithm based on the gradient boosting framework that constructs multiple additive trees to progressively optimize the model's objective function. In each iteration, XGBoost computes first-order and second-order gradient information based on the current model's residuals and generates new regression trees to minimize weighted loss, thereby achieving efficient modeling of complex nonlinear relationships. Therefore, our proposed DPF-GFD model integrates the advantages of GNN-based automatic feature extraction and XGBoost-based accurate predictive performance. Below, we briefly introduce the core principles of XGBoost and its objective function definition.

The hypothesis space of XGBoost is an ensemble of Classification and Regression Trees (CART) \cite{breiman1984cart}. Given $N$ samples in the dataset $D$, each with $F'$ dimensional features, by computing $K$ trees, the cumulative value of these $K$ trees is the prediction value. Therefore, the output of a node is:
\begin{equation}
    \hat{y}_i^{(K)} = \sum_{j=1}^K f_j(x_i) = f_1(x_i) + f_2(x_i) + \cdots + f_{k-1}(x_i) + f_k(x_i),
\end{equation}
where the model parameters are K trees, $\mathcal{F}$ = $\{ f_1, f_2, \cdots, f_K \}$.

The objective function of XGBoost includes a loss function component $L$ and a regularization component $\Omega$. The loss function measures the model's fit to the training data, while regularization measures the model's complexity. Therefore, the objective function is:
\begin{equation}
    \text{Obj}(\theta) = \sum_{i=1}^N L(y_i, \hat{y}_i) + \sum_{j=1}^K \Omega(f_j),
\end{equation}
where $L(y_i, \hat{y}_i)$ represents the loss function between sample label $ y_i $ and prediction $\hat{y}_i$, and $\Omega(f_j)$ is the regularization term for complexity to prevent overfitting.

Compared with traditional neural networks or linear classifiers, XGBoost has the following advantages: (1) its tree-based structure can naturally handle nonlinear features and feature interactions, making it suitable for capturing complex pattern associations in financial fraud; (2) through built-in regularization terms and pruning strategies, XGBoost effectively prevents overfitting; (3) its gradient boosting mechanism performs stably when handling severely imbalanced data distributions and can improve fraud sample detection capability by adjusting sample weights and loss function parameters; (4) the algorithm has high training efficiency and supports parallel computing, facilitating its application to large-scale financial transaction graphs. In summary, using the embedding representation H generated by the aforementioned graph neural network modules as input features and then utilizing XGBoost for final classification not only fully exploits the expressive power of graph representation learning but also leverages the discriminative advantages of ensemble learning, thereby achieving accurate identification and robust detection of fraud nodes.

\textbf{Discussion}. The proposed DPF-GFD framework presents several advantages. First, by explicitly modeling multi-frequency graph signals through a dual-path design, the framework decouples structural anomaly enhancement from feature-consistency reconstruction. Second, the frequency-complementary filtering strategy improves robustness under heterophily and relation camouflage. Third, the combination of graph filtering and tree-based classification contributes to stable training behavior and flexible decision boundaries. Nevertheless, several limitations remain. The model requires constructing a similarity-based kNN graph, which introduces additional computational overhead. Moreover, the current formulation assumes a static undirected graph structure and employs a concatenation-based fusion mechanism. Future work will explore dynamic and directed graph extensions, as well as adaptive fusion strategies to further enhance modeling flexibility.

\section{Experiments}  \label{sec: Experiments}

In this paper, we design supervised node classification experiments to evaluate the effectiveness of the DPF-GFD model in fraud node identification. The model was tested on four publicly available real-world financial fraud datasets. We first introduce the datasets used in the experiments, enumerate the comparison models, then describe the experimental setup, and finally provide a detailed analysis and explanation of the experimental results.

\subsection{Experimental Setup}

\textbf{Datasets}. We use four datasets from real financial scenarios to evaluate our proposed method, with statistics shown in Table \ref{statistics}. The four real datasets are the financial statement fraud dataset FDCompCN \cite{wu2023splitgnn}, the credit card fraud detection dataset FFSD \cite{xiang2023semi}, the Bitcoin transaction dataset Elliptic \cite{weber2019anti}, and the financial lending dataset DGraph \cite{huang2022dgraph}.

\begin{table}[ht]
    \centering
    \caption{Statistical information of the four datasets used for evaluation, including the total number of nodes, edges, feature dimensions, normal nodes, fraudulent nodes, and unknown nodes.}
    \small
    \label{statistics}
    \begin{tabular}{c|c|c|c|c}
    \hline
   \textbf{Dataset}  &  \textbf{FDCompCN} &  \textbf{FFSD} &  \textbf{Elliptic} &  \textbf{DGraph} \\
    \hline
    Nodes & 5317 & 77881 & 203769 & 3700550  \\
    Edges & 10059 & 860968 & 234355 & 4300999 \\
    Features & 57 & 7 & 166  & 17 \\
    Normals & 4758 & 24387 & 42019 & 1210092 \\
    Frauds & 559 & 5256 & 4545 & 15509 \\
    Unknowns & 0 & 48238 & 157205 & 2474949 \\
    \hline
    \end{tabular}
\end{table}

\textbf{FDCompCN} \cite{wu2023splitgnn} is a dataset for detecting financial statement fraud of Chinese companies, constructing a multi-relational graph based on suppliers, customers, shareholders, and financial information disclosed in Chinese company financial statements. Financial statement fraud includes seven types of violations disclosed by Chinese regulatory authorities, including inflated profits, inflated assets, false statements, delayed disclosure, omission of important information, fraudulent disclosure, and general accounting violations. Companies with more than three violations are labeled as fraudulent samples, while other companies are labeled as benign samples.

\textbf{FDCompCN} \cite{wu2023splitgnn} is a benchmark constructed to support the study of fraudulent financial reporting among Chinese enterprises. It represents each firm within a multi-relational graph derived from publicly disclosed information on upstream and downstream partners, ownership relations, and key financial indicators. The dataset reflects seven categories of misconduct recognized by Chinese regulators, such as overstating earnings or assets, providing inaccurate disclosures, delaying required reports, omitting material information, and other accounting irregularities. Firms recorded with more than three such infractions are designated as fraudulent instances, whereas those below this threshold are treated as non-fraudulent.

\textbf{FFSD} \cite{xiang2023semi} serves as a benchmark for semi-supervised detection of credit card fraud. The annotations are derived from incident records submitted by cardholders and subsequently validated by experts in the financial sector. A transaction is assigned a positive label when it is either flagged by the user or deemed fraudulent during expert assessment; all remaining transactions receive a negative label. The original data contains only 7 features: transaction code, transaction initiator code, transaction acceptor code, transaction amount, transaction location code, different transaction type codes, and labels. After processing and feature engineering, we ultimately generate 63 features.

\textbf{Elliptic} \cite{weber2019anti} is a dynamic transaction network constructed from Bitcoin transaction data. Nodes in the network represent Bitcoin transactions. Edges in the network represent the payment flow of Bitcoin, i.e., Bitcoin flowing from one transaction to another. The dataset contains over 230,000 directed edges. Elliptic has three types of nodes: normal nodes, fraud nodes, and unknown nodes. Nodes tagged as fraudulent correspond to transaction patterns linked with unlawful or harmful activities, including online swindles, malicious software operations, extremist groups, ransomware incidents, and pyramid-style schemes. Nodes designated as legitimate are connected to entities such as trading platforms, wallet-service providers, mining participants, and other compliant businesses. A substantial portion of nodes remains without labels, mirroring the practical difficulty of obtaining complete and reliable annotations in real-world scenarios.

\textbf{DGraph} \cite{huang2022dgraph} is a dynamic transaction network constructed from financial lending data. Nodes in the network represent financial lending users. The dataset contains over 3 million nodes and over 4 million edges, each edge associated with a timestamp indicating when the user filled in emergency contact information. DGraph has three types of nodes: normal nodes, fraud nodes, and unknown nodes. Normal nodes refer to users who have had borrowing behavior and repaid on time, while fraud nodes refer to users who have at least one overdue repayment. Unknown nodes are users who have not yet had borrowing behavior.

\textbf{Evaluation metrics}.Fraud identification is generally formulated as a two-class prediction task. Owing to the pronounced skew in node labels, we evaluate the models using Recall at K (Rec@K), the F1 score, the Area Under the Receiver Operating Characteristic Curve (AUC), and the Area Under the Precision–Recall Curve (AP). In our experimental setup, nodes associated with fraudulent behavior are considered positive instances, while regular nodes serve as negative instances.

Rec@K quantifies the proportion of truly fraudulent instances captured within the top-K transactions ranked by the model’s fraud likelihood scores. In practical fraud-screening scenarios, this metric reflects how effectively an auditor would identify genuine fraud cases when only the K highest-risk transactions are inspected. The F1 score represents the harmonic mean of Precision and Recall, combining them into a single indicator that balances both false positives and false negatives. Its value lies between 0 and 1, with higher scores indicating better alignment between the model’s ability to identify fraud and its ability to avoid incorrect alarms. AUC measures overall discriminative ability across all possible decision thresholds. It can be interpreted as the probability that the model assigns a higher score to a randomly selected fraudulent instance than to a randomly selected benign one, providing a threshold-independent assessment of ranking performance. AP summarizes the average precision achieved across varying recall levels. Compared to AUC, AP places greater emphasis on the model’s effectiveness in identifying positive (fraudulent) cases, especially under high-recall conditions that are critical in fraud detection applications.

Among these metrics, AUC captures global ranking quality but is less informative for top-K auditing scenarios, whereas Rec@K focuses exclusively on the highest-risk portion of the ranking. AP offers a balanced perspective by integrating both ranking quality and class imbalance sensitivity. In contrast, the F1-measure is threshold-dependent and therefore cannot characterize performance across the full range of decision thresholds. Consequently, AP is used as the primary criterion for tuning model performance in our experiments.

\textbf{Comparison methods}. To validate the effectiveness of the DPF-GFD model in graph-based financial fraud detection, we compare DPF-GFD with three categories of baselines: classical GNN models, including GCN \cite{kipf2016semi}, GraphSAGE \cite{hamilton2017inductive}, and GAT \cite{velickovic2018graph}; GNN-based spatial domain anomaly detection methods, including GAS \cite{li2019spam}, PC-GNN \cite{liu2021pick}, and PMP \cite{zhuo2024partitioning}; and GNN-based spectral domain anomaly detection methods, including BernNet \cite{he2021bernnet}, AMNet \cite{chai2022can}, and BWGNN \cite{tang2022rethinking}.

\begin{itemize}
    \item GCN \cite{kipf2016semi}: A classical graph method that uses convolution operations on graphs to propagate information from nodes to their neighbors, enabling the network to learn representations for each node based on its local neighborhood.

    \item GraphSAGE \cite{hamilton2017inductive}: A scalable inductive method that learns node representations by deriving feature information from sampled subsets of each node’s surrounding vertices and combining them through a designated aggregation function.

    \item GAT \cite{velickovic2018graph}: A GNN framework with attention mechanisms that prioritizes nodes during information aggregation, focusing on the most informative parts.

    \item GAS \cite{li2019spam}: A scalable spam detection method that applies GCN to heterogeneous graphs and applies KNN to application-specific graph structures.

    \item PC-GNN \cite{liu2021pick}: A GNN framework for fraud detection that samples balanced labels and employs a learnable distance function to select informative neighbors.

    \item PMP \cite{zhuo2024partitioning}: A message passing paradigm designed for fraud detection, where neighbors with different classes are aggregated through different node-specific aggregation functions during the neighbor aggregation phase.

    \item BernNet \cite{he2021bernnet}: Estimates any filter on the graph normalized Laplacian spectrum through a K-order Bernstein polynomial approximation and designs its spectral properties by setting the coefficients of Bernstein bases.

    \item AMNet \cite{chai2022can}: This model employs a series of BernNet layers to model information distributed across various spectral ranges, enabling it to learn representations that reflect both lower- and higher-frequency components in a data-adaptive manner.

    \item BWGNN \cite{tang2022rethinking}: This approach leverages Beta-based kernels to construct a set of localized filters in both spatial and spectral domains, allowing the network to highlight irregular patterns that manifest predominantly in the higher-frequency regions of graph signals.
    
\end{itemize}

\textbf{Implementation details}. This experiment is implemented based on PyTorch 2.1.1 and DGL 2.2.1 \cite{wang2019deep}. Each dataset is divided into three parts. For the Elliptic dataset, data from the first 23 time steps is used as the training set, data from time steps 24-34 as the validation set, and data from time steps 35-49 as the test set, strictly following chronological order to avoid future information leakage. For the FDCompCN and FFSD datasets, 70\% of labeled nodes are selected as the training set, 15\% as the validation set, and the remaining 15\% as the test set. For DGraph, we follow the official data split provided in the original paper to ensure a fair and reproducible comparison.

The kNN graph is constructed using DGL's KNNGraph module with shared-memory brute-force search. Only top-$k$ neighbor indices are stored, resulting in $\mathcal{O}(Nk)$ memory complexity. The construction is performed once as offline preprocessing. The similarity graph is constructed solely from node features in a standard transductive learning setting, and no label information from the validation or test sets is used during graph construction or model training.

The Beta wavelet operator is realized through an explicit Laplacian polynomial expansion rather than eigen-decomposition. Its spectral response is expressed using a Bernstein polynomial basis, and filtering is computed via iterative sparse Laplacian propagation in the vertex domain. The overall computational complexity is $\mathcal{O}(K|E|)$, where $K$ denotes the polynomial order and $|E|$ is the number of edges.

For baseline models, GCN, GraphSAGE, and GAT use implementations from open-source platforms \cite{wang2019deep}, while other baseline models use source code provided by their authors. To adapt to various baseline settings, we appropriately converted the data format for each baseline model. To ensure fairness in hyperparameter tuning, we use random search to optimize hyperparameters. In one trial for each dataset, we randomly select a set of hyperparameters from each model's predefined search space. The number of iterations for all models is set to 200.

\subsection{Performance Comparison}

Table \ref{tab:results} presents the experimental results of different models on four real-world financial fraud detection datasets (FDCompCN, FFSD, Elliptic, and DGraph), evaluated using Recall@K, F1, AUC, and AP. All results are reported as mean $\pm$ standard deviation over five independent runs with different random seeds. Overall, the proposed DPF-GFD demonstrates consistently strong performance across all datasets. Compared with the nine baseline models, DPF-GFD achieves the best or highly competitive results on most evaluation metrics while maintaining relatively low variance, indicating stable and reliable performance under different random initializations. Further analysis of the experimental findings is provided as follows.

\begin{table*}[t]
\centering
\footnotesize 
\caption{Performance comparison (Mean $\pm$ Std). The best results are highlighted in bold.}
\label{tab:results}
\renewcommand{\arraystretch}{1.2} 
\setlength{\tabcolsep}{1.8pt} 

\begin{tabular}{l|cccc|cccc}
\hline
\textbf{Models} & \multicolumn{4}{c|}{\textbf{FDCompCN}} & \multicolumn{4}{c}{\textbf{FFSD}} \\ 
\cline{2-9}
 & Rec@K & F1 & AUC & AP & Rec@K & F1 & AUC & AP \\ 
\hline
GCN & 0.1905$\pm$0.0084 & 0.2245$\pm$0.0202 & 0.6065$\pm$0.0318 & 0.1763$\pm$0.0087 & 0.3914$\pm$0.0511 & 0.3763$\pm$0.0487 & 0.6768$\pm$0.0284 & 0.4240$\pm$0.0330 \\
GraphSAGE & 0.1810$\pm$0.0308 & 0.2185$\pm$0.0257 & 0.6031$\pm$0.0434 & 0.1778$\pm$0.0154 & 0.2520$\pm$0.1162 & 0.2907$\pm$0.1712 & 0.5780$\pm$0.1004 & 0.2592$\pm$0.0787 \\
GAT & 0.1667$\pm$0.0304 & 0.1936$\pm$0.0140 & 0.5784$\pm$0.0311 & 0.1643$\pm$0.0210 & 0.4324$\pm$0.0091 & 0.4662$\pm$0.0119 & 0.6564$\pm$0.0054 & 0.3962$\pm$0.0213 \\
GAS & 0.1857$\pm$0.0136 & 0.2369$\pm$0.0128 & 0.6358$\pm$0.0175 & 0.1812$\pm$0.0097 & 0.4365$\pm$0.0224 & 0.4587$\pm$0.0159 & 0.7032$\pm$0.0233 & 0.4347$\pm$0.0556 \\
PCGNN & 0.3357$\pm$0.0155 & 0.3131$\pm$0.0468 & 0.7465$\pm$0.0155 & 0.3050$\pm$0.0189 & 0.3123$\pm$0.1185 & 0.3354$\pm$0.1446 & 0.6253$\pm$0.0464 & 0.3456$\pm$0.0988 \\
PMP & 0.1738$\pm$0.0181 & 0.1949$\pm$0.0302 & 0.5807$\pm$0.0382 & 0.1680$\pm$0.0215 & 0.4461$\pm$0.0333 & 0.4827$\pm$0.0491 & 0.6665$\pm$0.0224 & 0.3772$\pm$0.0498 \\
AMNet & 0.3143$\pm$0.0322 & 0.3122$\pm$0.0409 & 0.7314$\pm$0.0195 & 0.2887$\pm$0.0236 & 0.4256$\pm$0.0175 & 0.3656$\pm$0.0671 & 0.6623$\pm$0.0184 & 0.4630$\pm$0.0164 \\
BernNet & 0.2286$\pm$0.0284 & 0.2184$\pm$0.0172 & 0.6455$\pm$0.0362 & 0.2185$\pm$0.0170 & 0.4066$\pm$0.0569 & 0.3936$\pm$0.0597 & 0.6340$\pm$0.0420 & 0.3495$\pm$0.0338 \\
BWGNN & 0.2571$\pm$0.0299 & 0.2769$\pm$0.0348 & 0.6771$\pm$0.0240 & 0.2367$\pm$0.0156 & 0.4662$\pm$0.0283 & 0.3397$\pm$0.1895 & 0.7346$\pm$0.0285 & 0.5097$\pm$0.0760 \\
\textbf{DPF-GFD} & \textbf{0.4524$\pm$0.0084} & \textbf{0.4457$\pm$0.0090} & \textbf{0.7950$\pm$0.0112} & \textbf{0.4537$\pm$0.0165} & \textbf{0.6545$\pm$0.0039} & \textbf{0.6529$\pm$0.0067} & \textbf{0.9105$\pm$0.0018} & \textbf{0.7679$\pm$0.0026} \\
\hline
\textbf{Models} & \multicolumn{4}{c|}{\textbf{Elliptic}} & \multicolumn{4}{c}{\textbf{DGraph}} \\ 
\cline{2-9}
 & Rec@K & F1 & AUC & AP & Rec@K & F1 & AUC & AP \\ 
\hline
GCN & 0.4293$\pm$0.1181 & 0.3254$\pm$0.0799 & 0.8481$\pm$0.0191 & 0.3526$\pm$0.0968 & 0.0486$\pm$0.0012 & 0.0467$\pm$0.0004 & 0.7332$\pm$0.0031 & 0.0310$\pm$0.0002 \\
GraphSAGE & 0.6236$\pm$0.0501 & 0.3246$\pm$0.0295 & 0.8828$\pm$0.0156 & 0.6367$\pm$0.0951 & 0.0500$\pm$0.0016 & 0.0480$\pm$0.0004 & 0.7589$\pm$0.0036 & 0.0332$\pm$0.0005 \\
GAT & 0.4700$\pm$0.0851 & 0.3777$\pm$0.0681 & 0.8747$\pm$0.0147 & 0.3321$\pm$0.0532 & \textbf{0.0512$\pm$0.0023} & 0.0510$\pm$0.0021 & 0.7579$\pm$0.0045 & 0.0333$\pm$0.0006 \\
GAS & 0.5471$\pm$0.0801 & 0.4269$\pm$0.0831 & 0.8776$\pm$0.0103 & 0.4990$\pm$0.1188 & 0.0491$\pm$0.0023 & 0.0483$\pm$0.0013 & 0.7517$\pm$0.0051 & 0.0319$\pm$0.0005 \\
PCGNN & 0.4930$\pm$0.0888 & 0.3117$\pm$0.0254 & 0.8765$\pm$0.0115 & 0.3874$\pm$0.1127 & 0.0356$\pm$0.0066 & 0.0444$\pm$0.0017 & 0.7190$\pm$0.0129 & 0.0264$\pm$0.0014 \\
PMP & 0.5759$\pm$0.0258 & 0.3303$\pm$0.0149 & 0.8773$\pm$0.0073 & 0.4818$\pm$0.0571 & 0.0478$\pm$0.0022 & 0.0497$\pm$0.0012 & 0.7623$\pm$0.0079 & 0.0333$\pm$0.0016 \\
AMNet & 0.4132$\pm$0.0964 & 0.2681$\pm$0.0593 & 0.8457$\pm$0.0353 & 0.3137$\pm$0.0719 & 0.0452$\pm$0.0020 & 0.0450$\pm$0.0007 & 0.7317$\pm$0.0007 & 0.0286$\pm$0.0002 \\
BernNet & 0.4673$\pm$0.0805 & 0.2968$\pm$0.0298 & 0.8508$\pm$0.0180 & 0.3598$\pm$0.0709 & 0.0487$\pm$0.0025 & 0.0458$\pm$0.0013 & 0.7408$\pm$0.0016 & 0.0306$\pm$0.0004 \\
BWGNN & 0.5146$\pm$0.0905 & 0.3425$\pm$0.0957 & 0.8690$\pm$0.0097 & 0.3968$\pm$0.0681 & 0.0504$\pm$0.0031 & 0.0500$\pm$0.0005 & 0.7628$\pm$0.0016 & 0.0340$\pm$0.0005 \\
\textbf{DPF-GFD} & \textbf{0.7286$\pm$0.0055} & \textbf{0.6095$\pm$0.0930} & \textbf{0.9230$\pm$0.0061} & \textbf{0.7819$\pm$0.0045} & 
0.0489$\pm$0.0038 & \textbf{0.0694$\pm$0.0004} & \textbf{0.7637$\pm$0.0035} & \textbf{0.0347$\pm$0.0006} \\
\hline
\end{tabular}
\end{table*}

Compared with classical GNNs such as GCN, GraphSAGE, and GAT, DPF-GFD demonstrates consistent and stable improvements across the majority of evaluation metrics on the four datasets. Although classical GNNs can leverage mutual interactions between node features and graph structure for fraud detection, their effectiveness in complex fraud scenarios remains limited. This limitation is largely attributable to relational camouflage and severe class imbalance, which prevent fraud nodes from being sufficiently distinguished during the message-passing process. Traditional GNN aggregation mechanisms tend to smooth representations over neighborhoods, which may obscure high-frequency anomalous signals and weaken the separability between fraud and normal nodes.

By introducing dual-path filtering, DPF-GFD enhances structural discrimination while suppressing noisy propagation. The adaptive spectral filtering module captures multi-frequency structural characteristics from the original graph, while the similarity-graph low-pass filtering module reinforces feature consistency within reconstructed neighborhoods. The experimental results, reported as mean $\pm$ standard deviation across five independent runs, further confirm that the performance gains of DPF-GFD are consistent and robust rather than incidental, demonstrating improved stability across different random seeds.

Compared with spatial-domain anomaly detection methods such as GAS, PC-GNN, and PMP, DPF-GFD maintains stronger robustness across datasets with diverse structural properties. Spatial methods mainly rely on local neighborhood propagation and sampling strategies. However, in financial transaction networks where fraud nodes often exhibit cross-community connections or sparse structural patterns, purely local propagation mechanisms are vulnerable to relation camouflage and heterophily. Such mechanisms may lead to feature contamination and misclassification due to the mixing of fraudulent and benign neighborhood signals.

DPF-GFD addresses this issue through a dual-path filtering architecture. The adaptive filtering on the original graph enables multi-scale structural modeling, while the similarity-graph filtering reconstructs node relationships based on feature similarity rather than raw adjacency. This combination effectively alleviates the sensitivity of spatial models to structural noise. The consistently lower variance observed across repeated runs further indicates that DPF-GFD achieves not only improved effectiveness but also enhanced training stability.

Compared with spectral-domain methods such as AMNet, BernNet, and BWGNN, DPF-GFD also exhibits clear advantages. Spectral models typically assume that graph connectivity reflects semantic similarity. However, in fraud networks, malicious nodes deliberately create misleading connections to camouflage their behavior. Under such conditions, conventional spectral filtering may propagate and amplify mixed signals, reducing discriminative capacity.

DPF-GFD mitigates this limitation by introducing a similarity-based graph filtering branch that reconstructs neighborhood relationships in feature space. This design reduces the influence of camouflage connections and enhances the structural consistency of anomalous nodes. As demonstrated by the consistent improvements across multiple evaluation metrics and datasets, the dual-path filtering mechanism effectively balances anomaly preservation and noise suppression.

Overall, the experimental results indicate that DPF-GFD achieves strong overall performance among the comparison methods while maintaining stable variance across repeated trials. The effectiveness of the model stems from its dual-path filtering architecture and ensemble-based classification framework, which together enable more accurate modeling of complex fraud patterns under heterophily, camouflage, and severe class imbalance.

\subsection{Ablation Study}

The core components of the model are dual-path filtering on the original graph and neighbor graph and tree ensemble model classification. To verify the role of these three components in the model, we conduct ablation experiments by removing each component to assess their contributions to fraudster identification. The three variants tested are:

DPF-GFD\textbackslash O: Removes the adaptive filtering module based on the original graph, retaining only the low-pass filtering module based on the neighbor graph and the ensemble tree classification module.

DPF-GFD\textbackslash K: Removes the low-pass filtering module based on the neighbor graph, retaining only the adaptive filtering module based on the original graph and the ensemble tree classification module.

DPF-GFD\textbackslash T: Removes the tree ensemble module, using filtered node representations directly for classification.

\begin{figure*}[htbp]
    \centering
    \includegraphics[width=0.8 \textwidth]{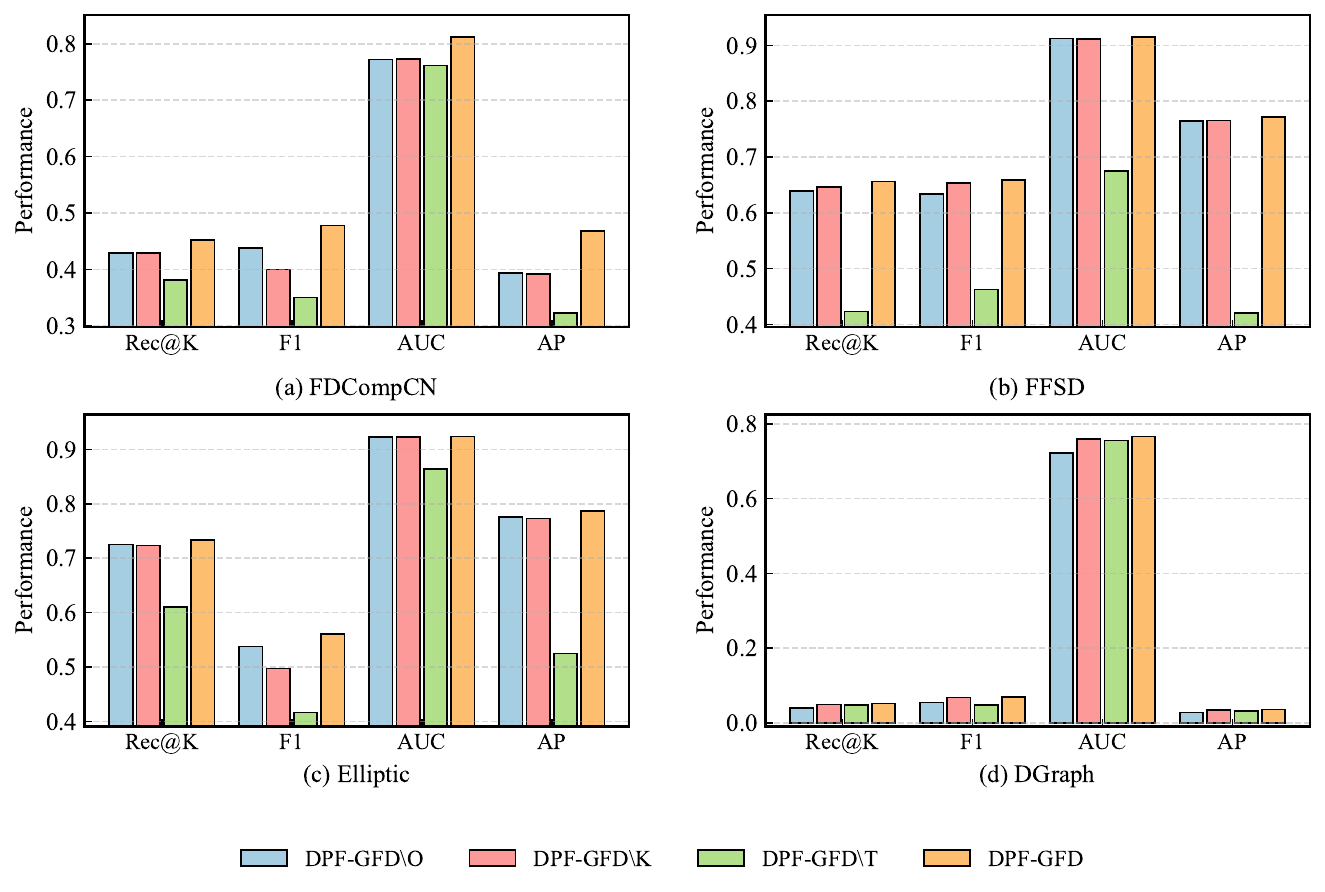}
    \caption{Results of the ablation study on four datasets.}
    \label{fig:ablation_study}
\end{figure*}

We compare these three model variants with the original model on four datasets. All models use the same parameters and experimental environment, with fixed random seeds to ensure fair comparison. Experimental results are shown in Figure \ref{fig:ablation_study}. Overall, the complete DPF-GFD model achieves optimal performance across all evaluation metrics, indicating that each module plays a positive role in improving the model's overall performance. Removing any module leads to performance degradation, though the impact varies across modules.

Specifically, removing the adaptive filtering module (DPF-GFD\textbackslash O) and removing the low-pass filtering module (DPF-GFD\textbackslash K) both affect the four performance metrics, indicating that these two modules play key roles in high-frequency feature extraction and low-frequency structural smoothing, respectively. After removing the adaptive filtering module (DPF-GFD\textbackslash O), the model's Recall@K and AUC metrics show significant declines, indicating that the adaptive filtering module plays a core role in capturing high-frequency anomalous signals. This module enhances node features in the frequency domain through a multi-frequency response mechanism, enabling the model to more accurately identify potential fraud samples in complex heterophilic relationship networks. In contrast, removing the low-pass filtering module (DPF-GFD\textbackslash K) mainly causes declines in F1 and AP metrics, indicating that the low-pass filtering module plays an important role in smoothing local features and reducing noise interference. This module weakens interference from camouflage connections through structural reconstruction and feature smoothing on neighbor subgraphs, making anomalous nodes more consistent at structural and feature levels, thereby improving the model's discriminative stability and feature expression quality in local neighborhoods. Furthermore, after removing the ensemble tree module (DPF-GFD\textbackslash T), the overall performance degradation is most significant, especially in F1 and AP metrics, indicating that the enhancement effect of the ensemble tree module at the discriminative layer is extremely significant. Since fraud detection data typically exhibit severe imbalance and noise, this module constructs more robust decision boundaries in the embedding space through ensemble learning mechanisms, effectively improving classification stability and generalization capability, thereby significantly enhancing the model's comprehensive performance.

In summary, the various components of the DPF-GFD model form complementary mechanisms in feature extraction and discrimination stages: the adaptive filtering module captures high-frequency anomalous features, strengthening the model's sensitivity to anomalous nodes; the low-pass filtering module achieves local structural smoothing and camouflage correction, improving feature consistency and noise resistance; and the ensemble tree module strengthens final discriminative performance and generalization capability. The synergistic effect of these three components enables DPF-GFD to achieve both high recall and high precision across various graph datasets, validating the rationality and effectiveness of the model design.

\subsection{Control Experiments with Fixed Classifier}
To investigate whether the performance gain mainly stems from the use of XGBoost, we conduct additional control experiments with a fixed XGBoost classifier. Specifically, we compare: (1) Raw Features + XGBoost, (2) GCN Embeddings + XGBoost, (3) DPF-GFD. All models are evaluated using AP (mean $\pm$ std over 5 runs).

\begin{table*}[t]
\centering
\small
\caption{Control experiments with a fixed XGBoost classifier. Results are reported in AP (mean $\pm$ std over 5 runs).}
\label{tab:control experiment}
\renewcommand{\arraystretch}{1.2}
\setlength{\tabcolsep}{9pt}
\begin{tabular}{l|cccc}
\hline
\textbf{Models} & \textbf{FDCompCN} & \textbf{FFSD} & \textbf{Elliptic} & \textbf{DGraph} \\
\hline
Raw + XGB 
& $0.4247 \pm 0.0153$ 
& $0.7497 \pm 0.0070$ 
& $0.7571 \pm 0.0035$ 
& $0.0271 \pm 0.0003$ \\

GCN Emb + XGB 
& $0.2163 \pm 0.0196$ 
& $0.6170 \pm 0.0095$ 
& $0.5052 \pm 0.0440$ 
& $0.0320 \pm 0.0003$ \\

DPF-GFD 
& $\mathbf{0.4537 \pm 0.0165}$ 
& $\mathbf{0.7679 \pm 0.0025}$ 
& $\mathbf{0.7819 \pm 0.0045}$ 
& $\mathbf{0.0347 \pm 0.0006}$ \\

\hline
\end{tabular}
\end{table*}

As shown in Table \ref{tab:control experiment}, although XGBoost improves performance over raw features, DPF-GFD consistently achieves the highest AP across all four datasets. Compared with Raw Features + XGBoost, DPF-GFD improves AP by $2\%-3\%$ on FDCompCN and Elliptic, and achieves more stable performance on FFSD with significantly lower variance. In the highly imbalanced DGraph dataset, although absolute AP values are low, DPF-GFD still provides consistent improvement over both raw features and standard GCN embeddings, demonstrating its effectiveness in large-scale, extreme-imbalance scenarios. These results confirm that the performance gain is not solely attributed to the use of XGBoost, but rather to the more discriminative node representations learned by the proposed dual-path filtering mechanism.

\subsection{Hyperparameter Analysis}

To comprehensively demonstrate the robustness of the proposed model, two groups of hyperparameter sensitivity experiments were conducted on four datasets: FDCompCN, FFSD, Elliptic, and DGraph. These experiments were designed to examine the influence of the filter parameter C and the neighborhood size parameter K used in graph construction on model performance.

\begin{figure*}[htbp]
    \centering
    \includegraphics[width=0.8 \textwidth]{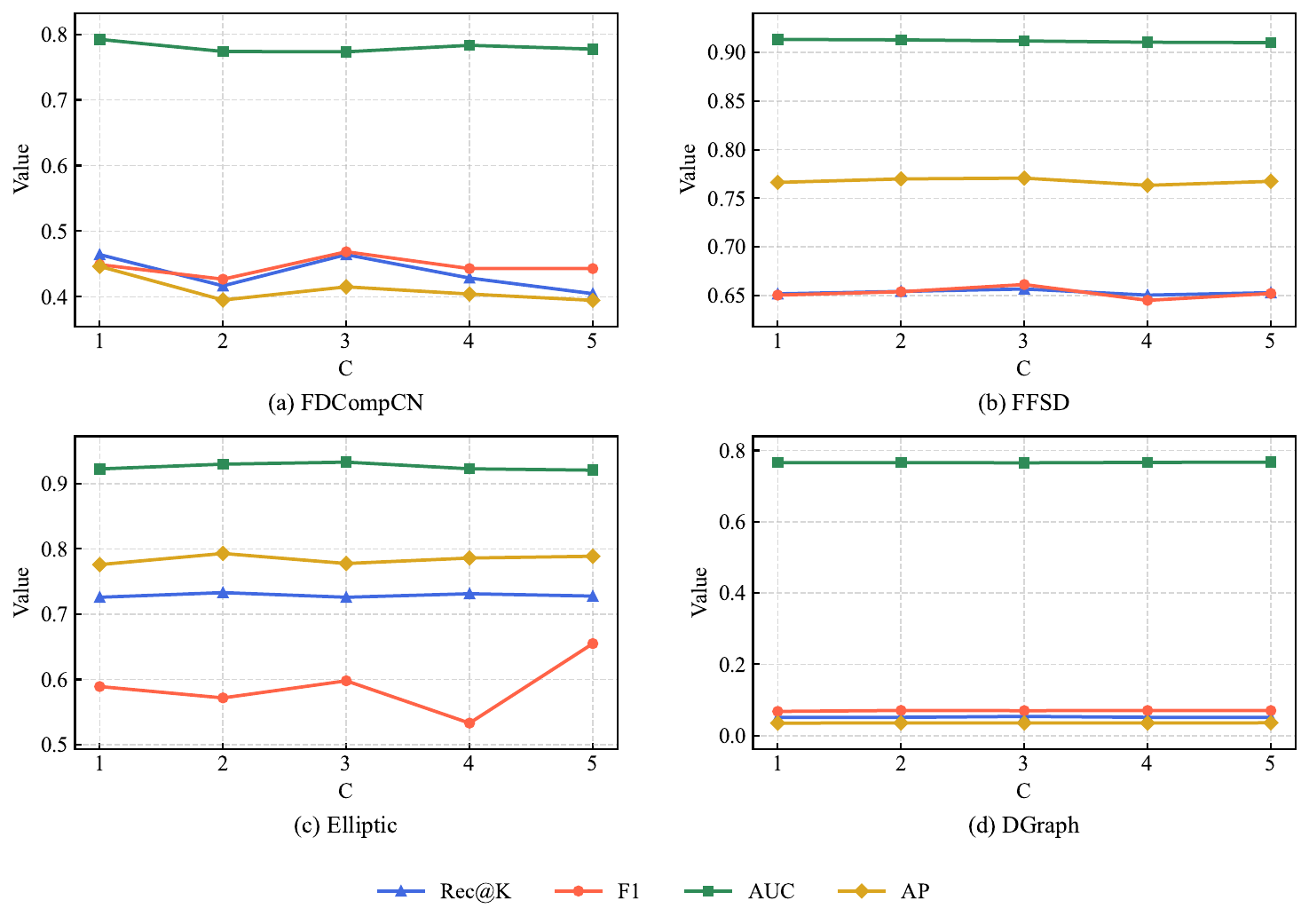}
    \caption{Performance under different values of parameter $C$ on four datasets.}
    \label{fig:hyperparam_C}
\end{figure*}

The Beta wavelet transform consists of a set of C+1 Beta wavelets with the same order; hence, C serves as a critical control parameter in the DPF-GFD model. As illustrated in Figure \ref{fig:hyperparam_C}, the sensitivity results with respect to C reveal that the performance curves across all four datasets exhibit strong consistency and stability. For FDCompCN, FFSD, and DGraph, all performance metrics (Rec@K, F1, AUC, and AP) remain almost constant, indicating that variations in C exert only a minor effect on the model. In the Elliptic dataset, a slight improvement in F1 is observed at higher orders (C = 5), while the ranking metrics (Rec@K, AUC, and AP) fluctuate only marginally. This trend suggests that increasing C introduces additional band-pass filters without significantly altering the model’s frequency response characteristics. In other words, the model at moderate orders is already capable of capturing the dominant spectral information of graph signals, and further increases in C yield diminishing returns. The marginal gain observed in Elliptic may result from an enhanced ability to decompose richer spectral components, but in most cases, higher orders provide limited benefit. Therefore, DPF-GFD demonstrates strong stability with respect to C, and choosing C = 2 or C = 3 provides an effective balance between computational efficiency and detection performance.

\begin{figure*}[htbp]
    \centering
    \includegraphics[width=0.8 \textwidth]{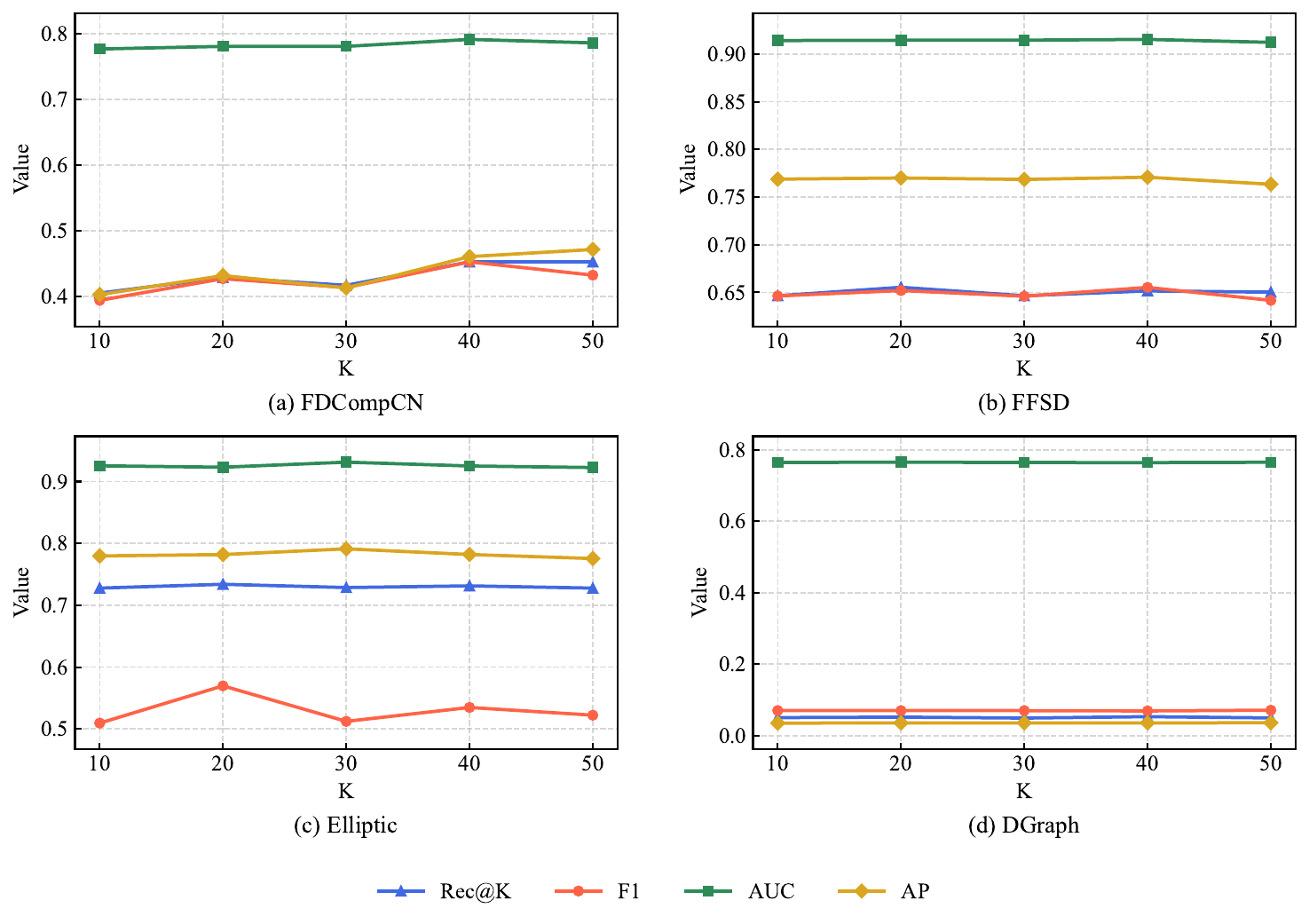}
    \caption{Performance under different values of parameter K on four datasets.}
    \label{fig:hyperparam_K}
\end{figure*}

The second group of experiments investigates the effect of the neighborhood size K in constructing the k-Nearest Neighbor graph. Specifically, K was set to 10, 20, 30, 40, and 50 under identical experimental configurations. As shown in Figure \ref{fig:hyperparam_K}, the performance variations across the four datasets remain small. For FDCompCN, FFSD, and DGraph, all four metrics fluctuate within a narrow range, indicating that the model is relatively insensitive to changes in K. The Elliptic dataset shows slightly larger fluctuations in F1 around K = 20. Overall, as K increases, the graph structure becomes denser, enabling more effective information propagation among nodes and leading to minor improvements in ranking metrics for certain datasets. However, when K becomes excessively large, irrelevant or noisy neighbors are incorporated, blurring the distinction between normal and anomalous nodes and causing a slight degradation in performance. Given that the overall variation remains minimal, DPF-GFD can be considered stable across a broad range of K values.

In summary, the sensitivity experiments on parameters C and K collectively demonstrate that DPF-GFD consistently maintains high and stable detection performance under varying hyperparameter configurations. The consistent trends observed across multiple datasets further confirm the model's strong generalization capability and robustness to hyperparameter variations.

\subsection{Visualization Analysis}m

\begin{figure*}[htbp]
    \centering
    \includegraphics[width=0.76 \textwidth]{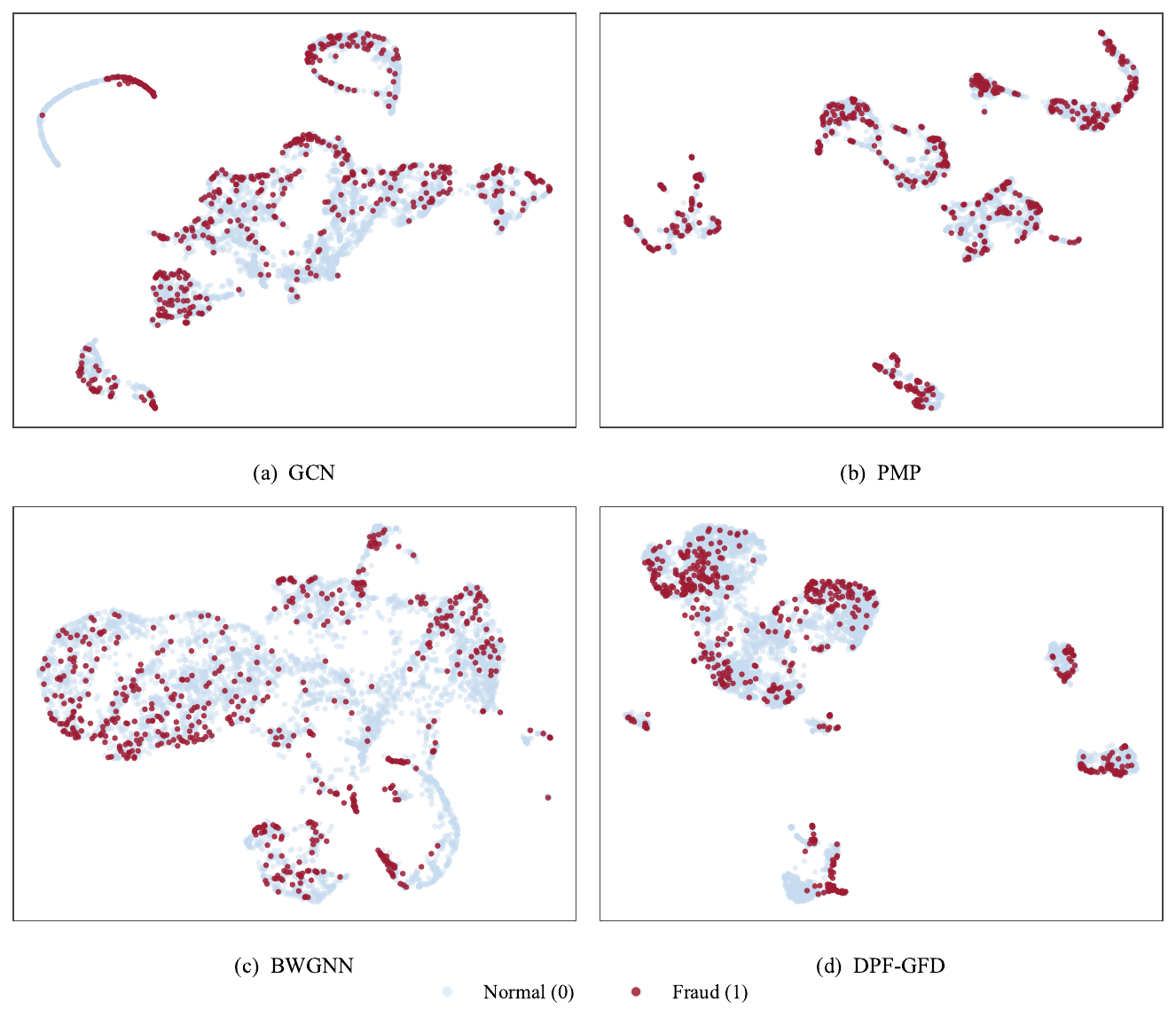}
    \caption{Visualization of the learned node embeddings on the FDCompCN dataset.}
    \label{fig:umap_binary}
\end{figure*}

To further intuitively demonstrate the effectiveness of our proposed model in node representation learning, we conduct a visualization analysis of GCN, PMP, BWGNN, and DPF-GFD on the FDCompCN dataset. Specifically, we extract node embeddings from the last layer of each model before classification and use the Uniform Manifold Approximation and Projection (UMAP) algorithm \cite{mcinnes2018umap} for dimensionality reduction and visualization of high-dimensional features. Figure \ref{fig:umap_binary}  shows the node distribution of four representative models in two-dimensional space, where normal nodes and fraud nodes are represented in light yellow and dark green, respectively.

From the figure, we can observe that the embedding results of GCN and PMP are relatively scattered, with significant overlap between classes, indicating that these two models struggle to fully capture node heterophily and local anomaly patterns when processing complex financial graphs. GCN shows a large number of mixed nodes at cluster boundaries, while PMP, although forming several aggregation clusters in local structure, still has a high overlap ratio between fraud samples and normal nodes, indicating that modeling methods relying solely on spatial neighborhood aggregation struggle to effectively distinguish anomalous nodes under camouflaged relationships. In contrast, BWGNN with spectral enhancement achieves some improvement in feature separability, with its node embeddings showing more obvious inter-class boundaries, but there are still some anomalous samples incorrectly absorbed into normal clusters. This indicates that relying solely on spectral filtering can alleviate feature smoothing problems, but it still struggles to achieve stable anomaly differentiation when facing relation camouflage and edge perturbations.

The visualization results of DPF-GFD further demonstrate a more organized embedding structure. We observe that although normal nodes and fraud nodes are generally located in the same cluster, they show a clear differentiation trend in spatial distribution: normal nodes are mainly concentrated on one side of the cluster, while fraud nodes increasingly aggregate on the other side, forming a continuous but directionally distinguishable distribution pattern. This co-cluster differentiation phenomenon indicates that DPF-GFD, while maintaining global spatial consistency, can learn potential class distinction directions within the same feature manifold, achieving implicit discrimination of camouflaged relationships. This property benefits from our proposed dual-filtering feature modeling mechanism: the adaptive filtering module captures multi-frequency features in the spectral domain, enhancing the model's response to high-frequency anomalous signals; the low-pass filtering module reconstructs neighborhood relationships at the structural level, effectively weakening the interference of camouflage edges. The synergistic effect of both enables DPF-GFD to achieve natural separation of feature directions in the shared embedding space, thus showing higher discriminability and structural consistency in node embedding distribution.

In summary, UMAP visualization results intuitively validate the advantages of DPF-GFD in the node representation learning stage. Compared with other baseline models, the node embeddings learned by DPF-GFD exhibit smoother class transitions and stronger directional differentiation features within the same cluster, indicating that the model can effectively capture potential semantic differences between fraud samples and normal nodes, providing a more robust representation foundation for subsequent classification and detection tasks.

\section{Conclusion} \label{sec: conclusion}

We propose the DPF-GFD model to address the challenges of relation camouflage, heterophily, and class imbalance in financial fraud detection. The DPF-GFD model consists of four components: an adaptive filter for the original graph, a low-pass filter for the neighbor graph,  a feature fusion module, and a final ensemble tree classifier. The adaptive filtering module is well-suited for fraud detection tasks with severe class imbalance and enables the trained filter to capture specific and appropriate signal frequencies. Using original features to construct a k-nearest neighbor graph captures the underlying structure of nodes in feature space, which effectively overcomes the erroneous connections introduced by fraud nodes. For the constructed neighbor graph, the model applies a low-pass filtering component to adaptively capture similar feature patterns. Finally, through ensemble tree classification, the model integrates the advantages of GNN-based automatic feature extraction and XGBoost-based accurate predictive performance. Experiments on four publicly available datasets (FDCompCN, FFSD, Elliptic, and DGraph) validate the effectiveness of the DPF-GFD model. In future work, we plan to adapt our fraud-detection framework to settings where labeled nodes are scarce but unlabeled instances are plentiful. Such an extension would enable the model to uncover emerging fraudulent behaviors even when only a small portion of the graph has supervisory information. Specifically, we plan to explore semi-supervised and self-supervised learning frameworks to fully mine the potential structural information of unlabeled nodes; simultaneously, we will introduce representation learning and pseudo-label generation mechanisms to improve the model's generalization and stability in low-supervision environments. Additionally, we will study dynamic graph and temporal feature modeling methods to adapt to the temporal evolution of fraudulent behavior. Through these improvements, we expect to develop a more robust and scalable graph neural network-based fraud detection system for real-world financial scenarios.

\section*{Acknowledgment}
This research was supported in part by the National Natural Science Foundation of China (No. 62272196), the Guangzhou Basic and Applied Basic Research Foundation (No. 2024A04J9971).

\section*{Data Availability}
Data and code are available at https://github.com/vidahee/
DPF-GFD.

\section*{CRediT Authorship Contribution Statement}
\textbf{Wei He}: Methodology, writing original draft.
\textbf{Wensheng Gan}: Review and editing, Supervision.
\textbf{Philip S. Yu}: Review and editing.


\bibliographystyle{unsrt}

\bibliography{main.bib}

\end{document}